\documentclass{article}
\usepackage{spconf,amsmath,graphicx,lipsum}
\usepackage{xcolor}
\usepackage{float}
\usepackage{subcaption}
\usepackage{hyperref}

\DeclareMathOperator*{\argmax}{argmax}

\newcommand{\dataset}{NuScenes }
\newcommand{\datasetF}{NS-F }
\newcommand{\datasetFB}{NS-FB }
\newcommand{\ea}{\textit{et al. }}

\title{RRPN: Radar Region Proposal Network \\for Object Detection in Autonomous Vehicles}
\name{Ramin Nabati, Hairong Qi}
\address{Department of Electrical Engineering and Computer Science\\
The University of Tennessee Knoxville, USA}
\begin{document}
\maketitle
\begin{abstract}
Region proposal algorithms play an important role in most state-of-the-art 
two-stage object detection networks by hypothesizing object locations in 
the image. Nonetheless, region proposal algorithms are known to be the bottleneck 
in most two-stage object detection networks, increasing the processing time for each 
image and resulting in slow networks not suitable for real-time applications 
such as autonomous driving vehicles.
In this paper we introduce RRPN, a Radar-based real-time region proposal algorithm 
for object detection in autonomous driving vehicles. RRPN generates object 
proposals by mapping Radar detections to the image coordinate system and 
generating pre-defined anchor boxes for each mapped Radar detection point. These 
anchor boxes are then transformed and scaled based on the object's distance from 
the vehicle, to provide more accurate proposals for the detected objects.
We evaluate our method on the newly 
released \dataset dataset \cite{caesar2019nuscenes} using the Fast R-CNN object detection network 
\cite{Girshick2015}. Compared to the Selective Search object proposal 
algorithm \cite{ss2013}, our model operates more than 100$\times$ 
faster while at the same time achieves higher detection precision and recall.
Code has been made publicly available at \url{https://github.com/mrnabati/RRPN}.
\end{abstract}

\section{Introduction}
\label{sec:intro}

Real-time object detection is one of the most challenging problems in building 
perception systems for autonomous vehicles. Most self-driving vehicles take 
advantage of several sensors such as cameras, Radars and LIDARs. Having different 
types of sensors provides an advantage in tasks such as object detection and may 
result in more accurate and reliable detections, but at the same time makes 
designing a real-time perception 
system more challenging.

Radars are one of the most popular sensors used in autonomous vehicles and have been 
studied for a long time in different automotive applications. Authors in 
\cite{Grimes1974} were among the first researchers 
discussing such applications for Radars, providing a detailed approach for utilizing them 
on vehicles. While Radars can provide accurate range and range-rate information on the 
detected objects, they are not suitable for tasks such as object classification. 
Cameras on the other hand, are very effective sensors 
for object classification, making Radar and camera sensor fusion a very 
interesting topic in autonomous driving applications. Unfortunately, there has been 
very few studies in this area in recent years, mostly due to the lack of a publicly available 
dataset with annotated and synchronized camera and Radar data in an autonomous driving 
setting.

2D object detection has seen a significant progress over the past few years, resulting in 
very accurate and efficient algorithms mostly based on convolutional neural networks 
\cite{Girshick2015,rfcn2016r,ren2015faster,liu2016ssd}. These methods usually fall under two 
main categories, one-stage and two-stage algorithms. One-stage algorithms treat object 
detection as a regression problem and learn the class probabilities and bounding boxes 
directly from the input image \cite{soviany2018optimizing}. YOLO \cite{redmon2016YOLO} and 
SSD \cite{liu2016ssd} are among the most popular algorithms in this category. Two-stage 
algorithms such as \cite{Girshick2015,ren2015faster} on the other hand, use a Region Proposal 
Network (RPN) in the first stage to generate regions of interests, and then use these 
proposals in the second stage to do classification and bounding box regression. 
One-stage algorithms usually reach lower accuracy rates, but are much faster than 
their two-stage counterparts. The bottleneck in two-stage 
algorithms is usually the RPN, processing every single image to generate ROIs for the object classifier, although yielding higher accuracy. This 
makes two-stage object detection algorithms not suitable for applications such as autonomous driving where 
it's extremely important for the perception system to operate in real time.

In this paper we propose Radar Region Proposal Network (RRPN), a real-time RPN based on Radar detections in autonomous 
vehicles. 
By relying only on Radar detections to propose regions of interest, we bypass the 
computationally expensive vision-based region proposal step, while improving detection 
accuracy. We demonstrate the effectiveness of our approach in the newly released 
\dataset dataset \cite{caesar2019nuscenes}, featuring data from Radars and cameras among 
other sensors integrated on a vehicle. When used in the Fast R-CNN object detection network, 
our proposed method achieves higher mean Average Precision (AP) and mean Average Recall 
(AR) compared to the Selective Search algorithm originally used in Fast R-CNN, while 
operating more than 100$\times$ faster.
\section{Related Work}
\label{sec:related}
Authors in \cite{Gibson1994} discussed the application of Radars in navigation
for autonomous vehicles, using an extended Kalman filter to fuse the radar and
vehicle control signals for estimating vehicle position. In \cite{Miyahara2006} 
authors proposed a correlation based pattern matching algorithm in addition to 
a range-window to detect and track objects in front of a vehicle. In \cite{Ji2008} 
Ji \ea proposed an attention selection system based on Radar detections to find 
candidate targets and employ a classification network to classify those objects. 
They generated a single attention window for each Radar detection, and used a 
Multi-layer In-place Learning Network (MILN) as the classifier.

Authors in \cite{Premebida2009} proposed a LIDAR and vision-based pedestrian detection 
system using both a centralized and decentralized fusion architecture. 
In the former, authors proposed a feature level fusion system where features from LIDAR and 
vision spaces are combined in a single vector which is classified using a single 
classifier. In the latter, two classifiers are employed, one per sensor‐feature 
space. More recently, Choi \ea in \cite{Cho2014} proposed a multi-sensor fusion system 
addressing the fusion of 14 sensors integrated on a vehicle. This system uses an 
Extended Kalman Filter to process the observations from individual sensors and is 
able to detect and track pedestrian, bicyclists and vehicles.

Vision based object proposal algorithms have been very popular among object detection 
networks. Authors in \cite{ss2013} proposed the Selective Search algorithm, diversifying the search for objects by using a
variety of complementary image partitionings. Despite its high accuracy, Selective 
Search is computationally expensive, operating at 2-7 seconds per image. 
Edge Boxes \cite{edge2014} is another vision based object proposal algorithm 
using edges to detect 
objects. Edge Boxes is faster than the Selective Search algorithm with a run time of $0.25$ 
seconds per image, but it is still considered very slow in real-time 
applications such as autonomous driving.

\section{Radar Region Proposal Network}
\label{sec:rrpn}

\begin{figure*}[t]
\centering
 \begin{footnotesize}
  \begin{subfigure}{0.19\linewidth}
    \includegraphics[width=0.98\textwidth]{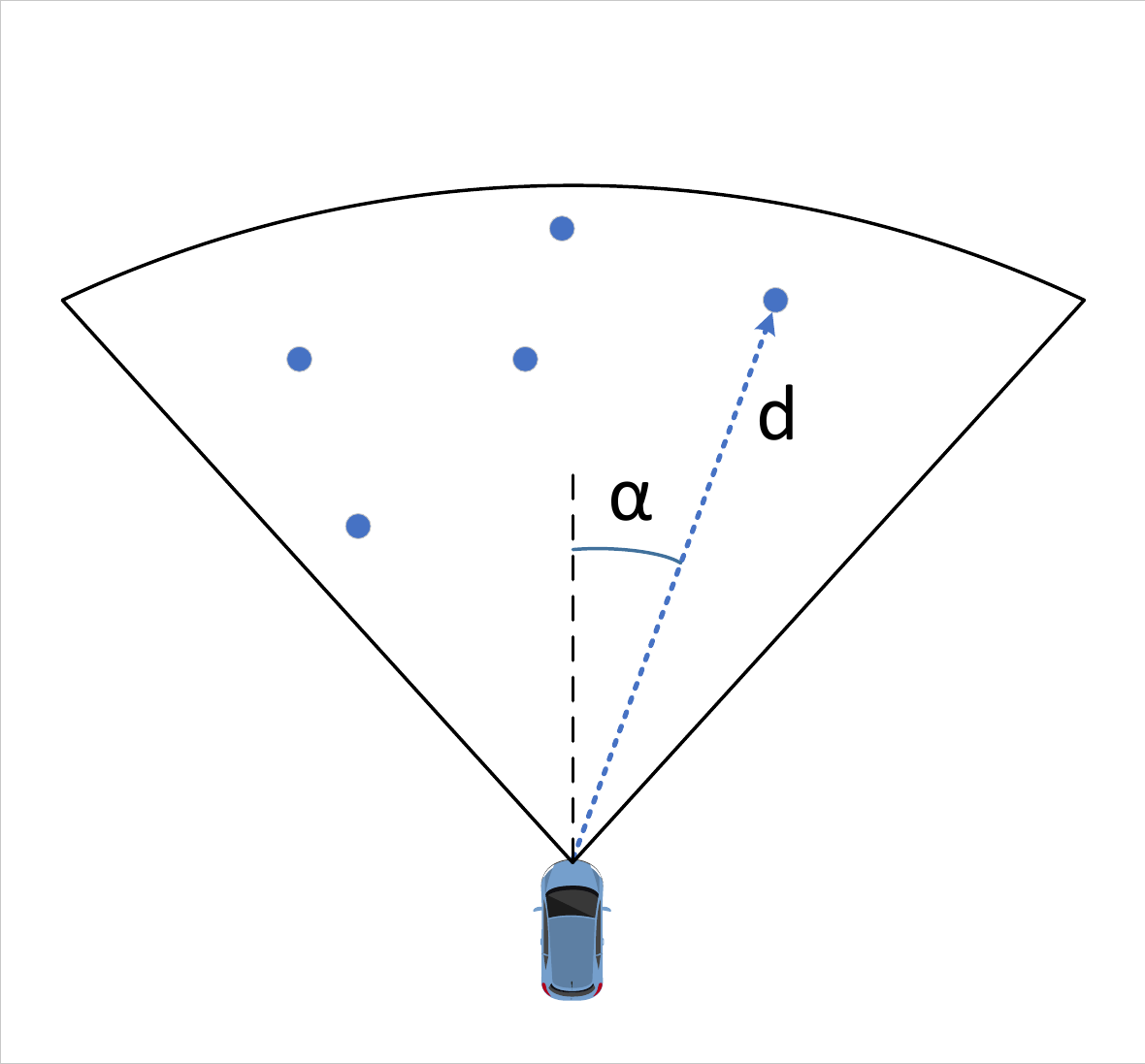}
    \caption{Bird's eye view}
  \end{subfigure}
  \begin{subfigure}{0.19\linewidth}
    \includegraphics[width=0.91\textwidth]{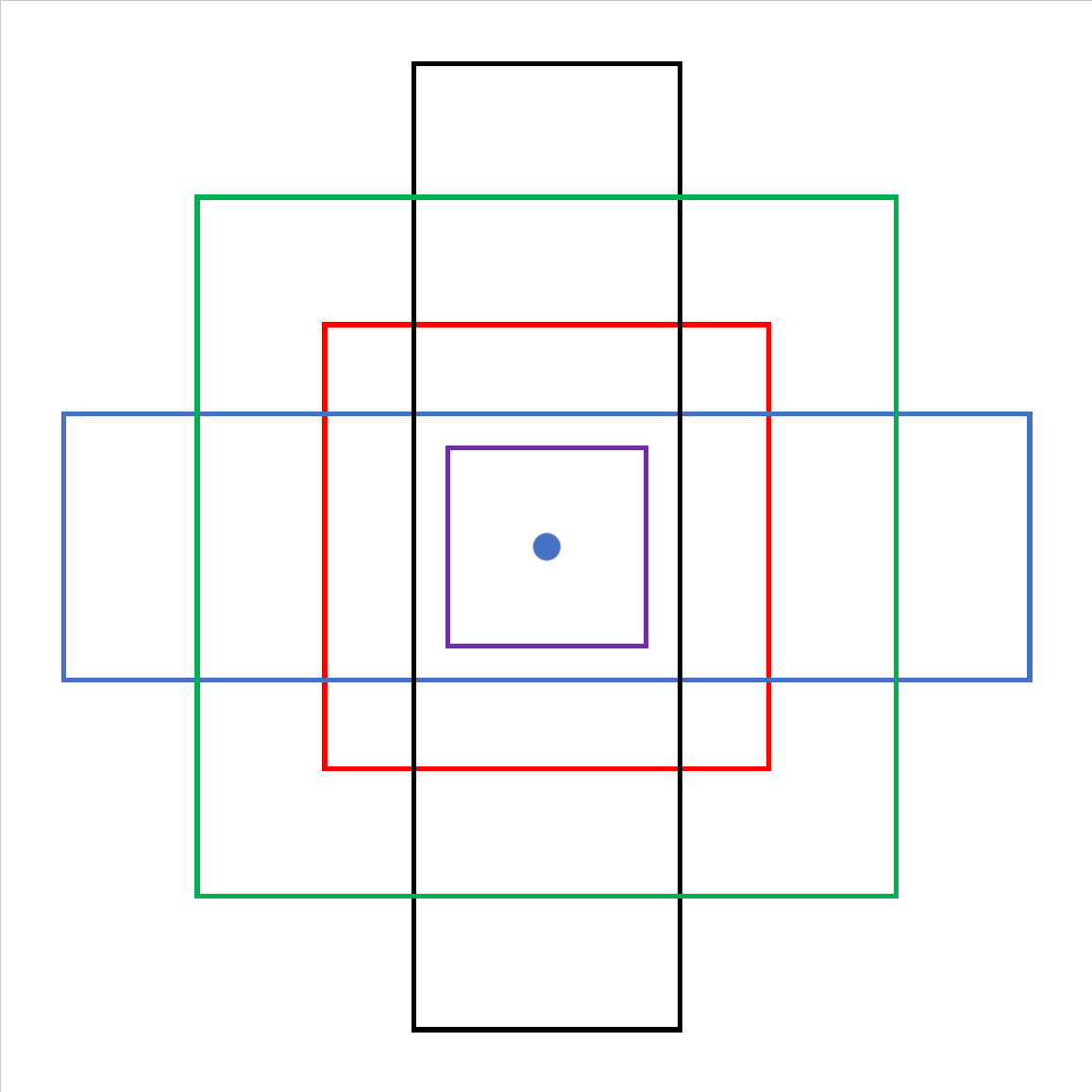}
    \caption{Centered anchors}
  \end{subfigure}
  \begin{subfigure}{0.19\linewidth}
    \includegraphics[width=0.91\textwidth]{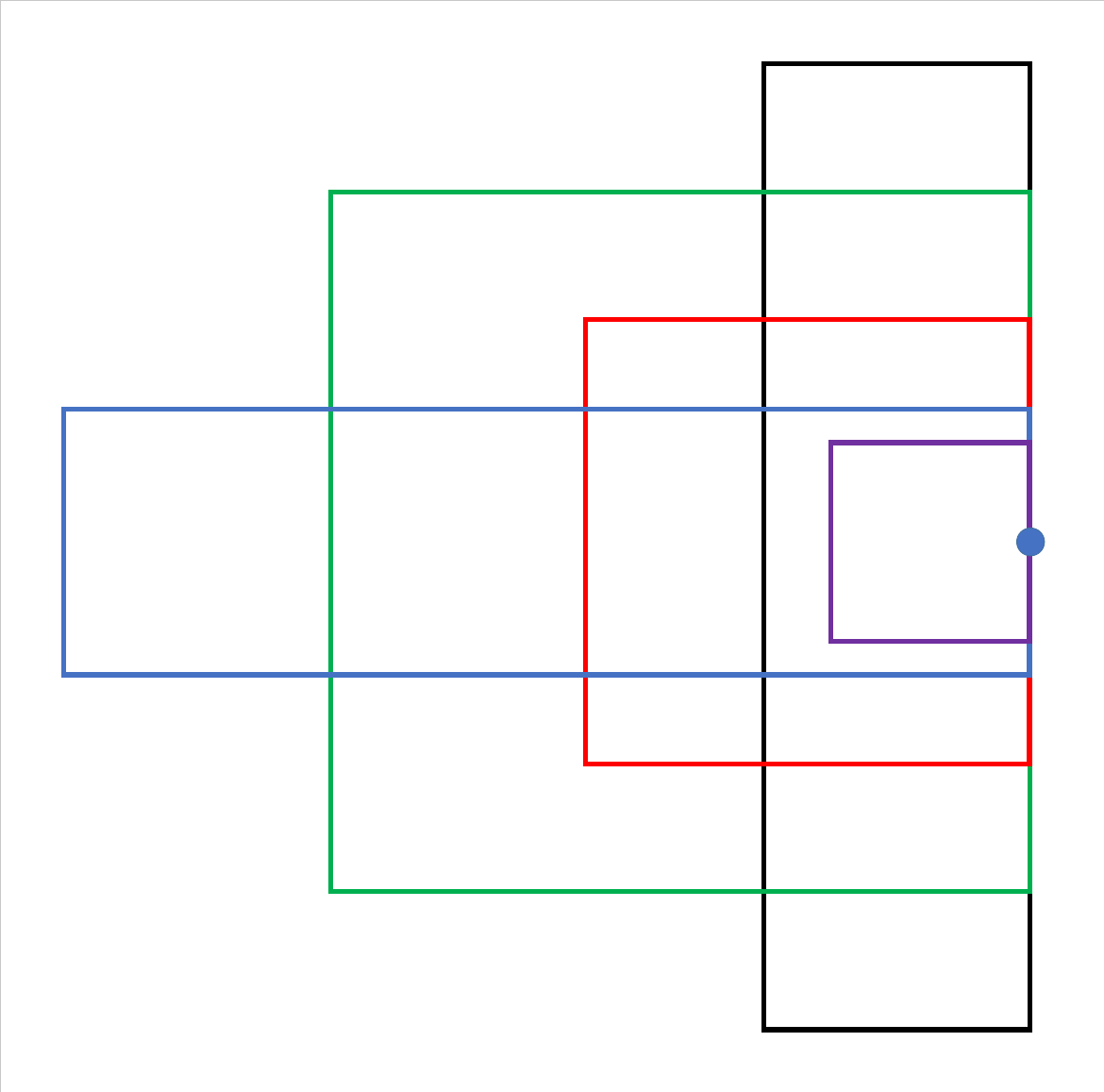}
    \caption{Right aligned anchors}
  \end{subfigure}
  \begin{subfigure}{0.2\linewidth}
    \includegraphics[width=0.86\textwidth]{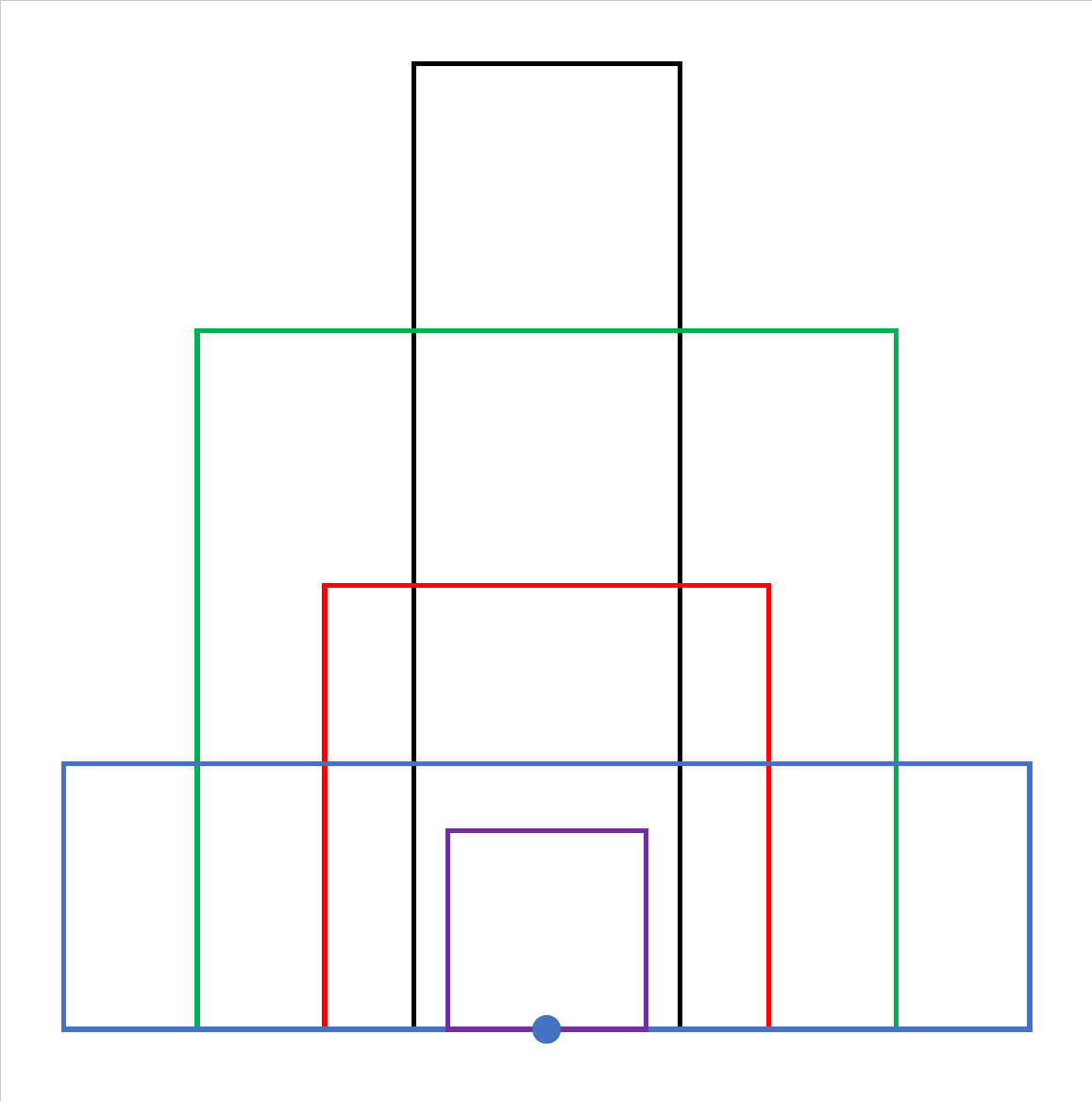}
    \caption{Bottom-aligned anchors}
  \end{subfigure}
  \begin{subfigure}{0.19\linewidth}
    \includegraphics[width=0.91\textwidth]{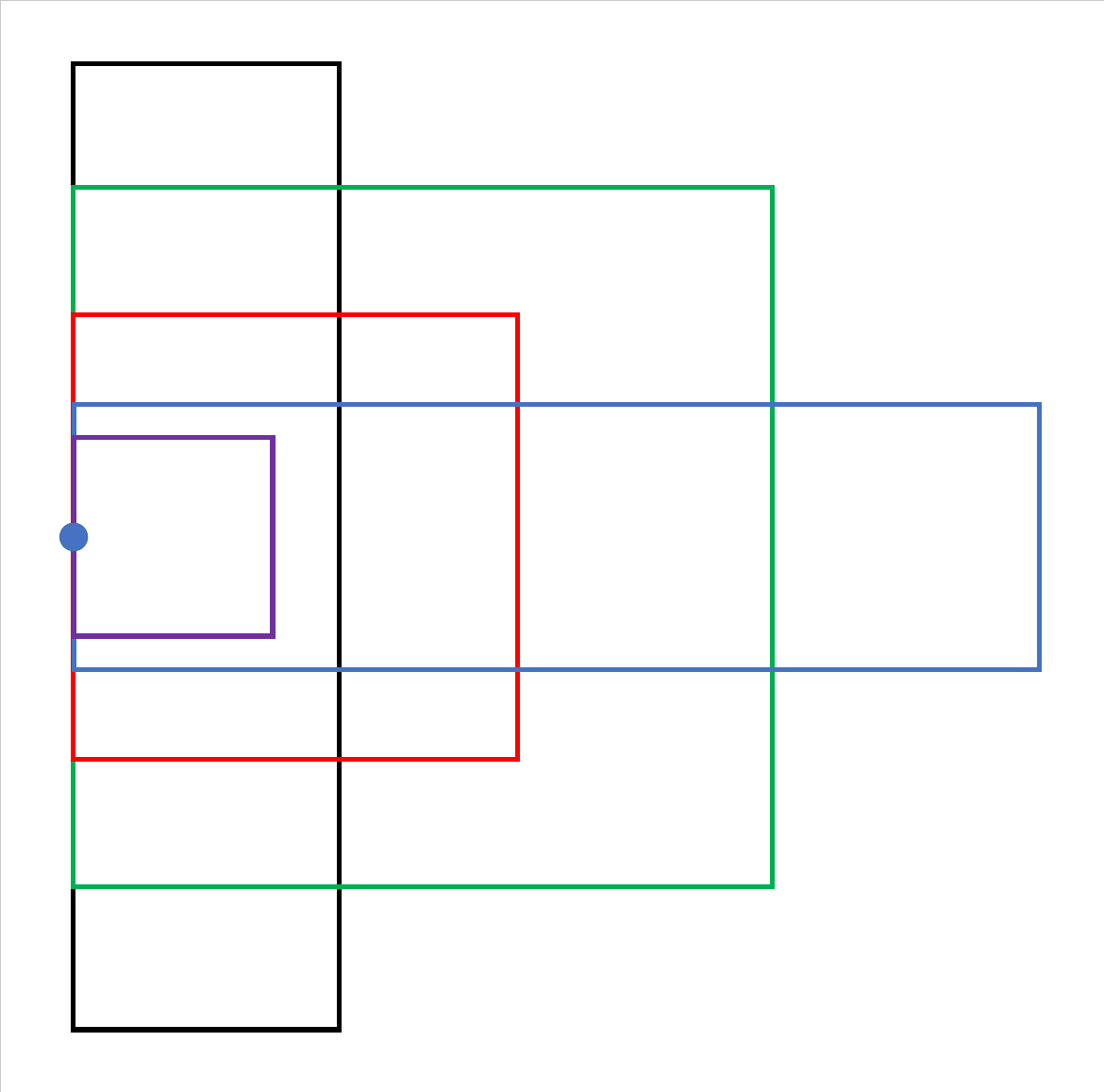}
    \caption{Left aligned anchors}
  \end{subfigure}
   \end{footnotesize}
  \caption{Generating anchors of different shapes and sizes for each Radar detection, shown here as the blue circle.}
  \label{fig:boxes}
\end{figure*}

We propose RRPN for object detection and classification in autonomous vehicles, a real-time algorithm using Radar detections 
to generate object proposals. The generated proposals can be used in 
any two-stage object detection network such as Fast-RCNN. Relying only on Radar 
detections to generate object proposals makes an extremely fast RPN, 
making it suitable for autonomous driving applications.
Aside from being a RPN for an object detection algorithm, the proposed network also 
inherently acts as a sensor fusion algorithm by fusing the Radar and camera data to 
obtain higher accuracy and reliability. The objects' range and range-rate information 
obtained from the Radar can be easily associated with the proposed regions of interest, 
providing accurate depth and velocity information for the detected objects.

RRPN also provides an attention mechanism to focus the underlying computational resources 
on the more important parts of the input data. While in other object detection 
applications the entire image may be of equal importance. In an autonomous driving 
application more attention needs to be given to objects on the road. 
For example in a highway driving scenario, the perception system needs to be able 
to detect all the vehicles on the road, but there is no need to dedicate resources to 
detect a picture of a vehicle on a billboard. A Radar based RPN 
focuses only on the physical objects surrounding the vehicle, hence inherently 
creating an attention mechanism focusing on parts of the input image that 
are more important.

The proposed RRPN consists of three steps: perspective transformation, anchor generation 
and distance compensation, each individually discussed in the following 
sections.

\subsection{Perspective Transformation}
\label{sec:persTrans}
The first step in generating ROIs is mapping the radar 
detections from the vehicle coordinates to the camera-view coordinates. Radar 
detections are reported in a bird's eye view perspective as shown in 
Fig. \ref{fig:boxes} (a), with the object's range and azimuth measured in the 
vehicle's coordinate system. By mapping these detections to the camera-view coordinates, 
we are able to associate the objects detected by the Radars to those seen in the 
images obtained by the camera.

In general, the projective relation between a 3D point $P =[X; Y; Z; 1]$ and its image
$p =[x; y; 1]$ in the camera-view plane can be expressed as below:
\begin{gather}
  p = HP  \hspace{3pt}, \hspace{30pt}
  H = \begin{bmatrix}
    h_{11} & h_{12} & h_{13} & h_{14} \\
    h_{21} & h_{22} & h_{23} & h_{24}\\
    h_{31} & h_{32} & h_{33} & h_{34}
  \end{bmatrix}
\label{eq:Transform}
\end{gather}

In an autonomous driving application, the matrix $H$ can be obtained from the calibration 
parameters of the camera.

\subsection{Anchor Generation}
Once the Radar detections are mapped to the image coordinates, we have the 
approximate location of every detected object in the image. These mapped Radar 
detections, hereafter called Points of Interest (POI), provide valuable 
information about the objects in each image, without any processing on the image 
itself. Having this information, a simple approach for proposing ROIs 
would be introducing a bounding box centered at every POI. One problem with this 
approach is that Radar detections are not always mapped to the center of the 
detected objects in every image. Another problem is the fact that Radars do not provide 
any information about the size of the detected objects and proposing a fixed-size 
bounding box for objects of different sizes would not be an effective approach.

We use the idea of anchor bounding boxes from 
Faster R-CNN \cite{ren2015faster} to alleviate the problems mentioned above. For every 
POI, we generate several bounding boxes with different sizes and aspect ratios 
centered at the POI, as shown in Fig.~\ref{fig:boxes} (b). 
We use 4 different sizes and 3 different aspect ratios to generate these anchors.

To account for the fact that the POI is not always mapped to the center of the object 
in the image coordinate, we also generate different translated versions of the 
anchors. These translated anchors provide more accurate bounding boxes when the 
POI is mapped towards the right, left or the bottom of the object as shown in 
Fig. \ref{fig:boxes} c-e.

\subsection{Distance Compensation}
The distance of each object from the vehicle plays an important role in 
determining its size in the image. Generally, objects' sizes in an image have 
an inverse relationship with their distance from the camera. Radar detections have 
the range information for every detected object, which is used in this step to 
scale all generated anchors. We use the following formula to determine the scaling factor 
to use on the anchors:
\begin{gather}
  S_i = \alpha \frac{1}{d_i} + \beta
\label{eq:scale}
\end{gather}
where $d_i$ is the distance to the $i$th object, and $\alpha$ and $\beta$ are two 
parameters used to adjust the scale 
factor. These parameters are learned by maximizing the Intersection Over Union (IOU) 
between the generated bounding 
boxes and the ground truth bounding boxes in each image, as shown in 
Eq. \ref{eq:maxIOU} below.
\begin{gather}
  \argmax_{\alpha, \beta} \hspace{8pt} \sum_{i=1}^{N} \sum_{j=1}^{M_i} \max_{1<k<A_i} IOU_{jk}^i(\alpha, \beta)
\label{eq:maxIOU}
\end{gather}
In this equation, $N$ is the number of training images, $M_i$ is the number of ground truth 
bounding boxes in image $i$, 
$A_i$ is the number of anchors generated in image $i$, and $IOU_{jk}^i$ is the IOU 
between the $j$th ground truth bounding box in image $i$ and the $k$th proposed 
anchor in that image. This equation finds the parameters $\alpha$ and $\beta$ 
that maximize the IOU between the ground truth and proposed bounding boxes. We use a simple grid search approach over a range of values to find $\alpha$ and $\beta$.

\section{Experiments and Results}
\label{sec:exp}
\begin{figure*}[t]
\centering
  \begin{minipage}{0.24\linewidth}
    \includegraphics[width=0.98\textwidth,trim={0 0.7cm 1.2cm 0},clip]{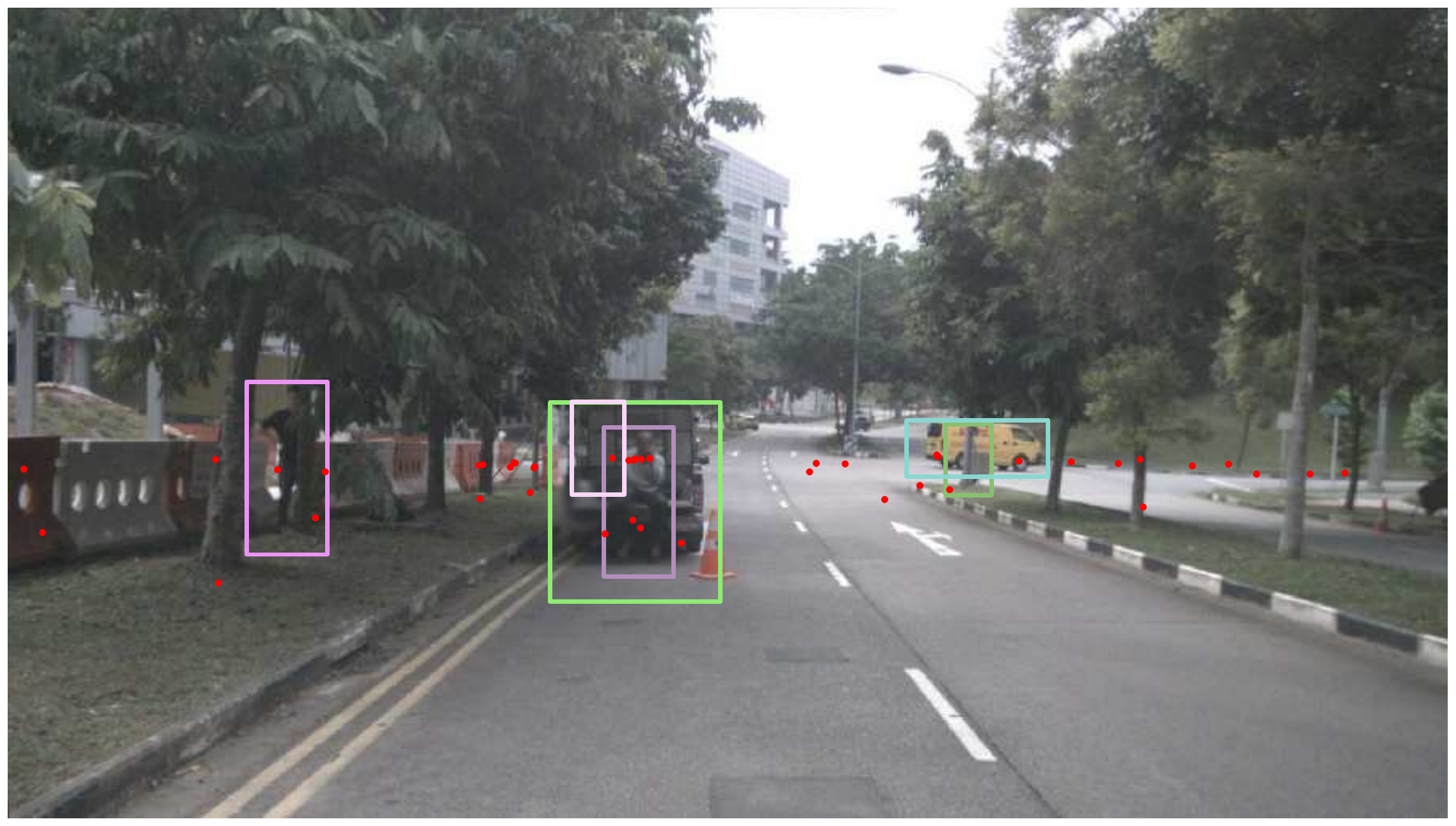}
  \end{minipage}
  \begin{minipage}{0.24\linewidth}
    \includegraphics[width=0.98\textwidth,trim={0 0.1cm 1.2cm 0cm},clip]{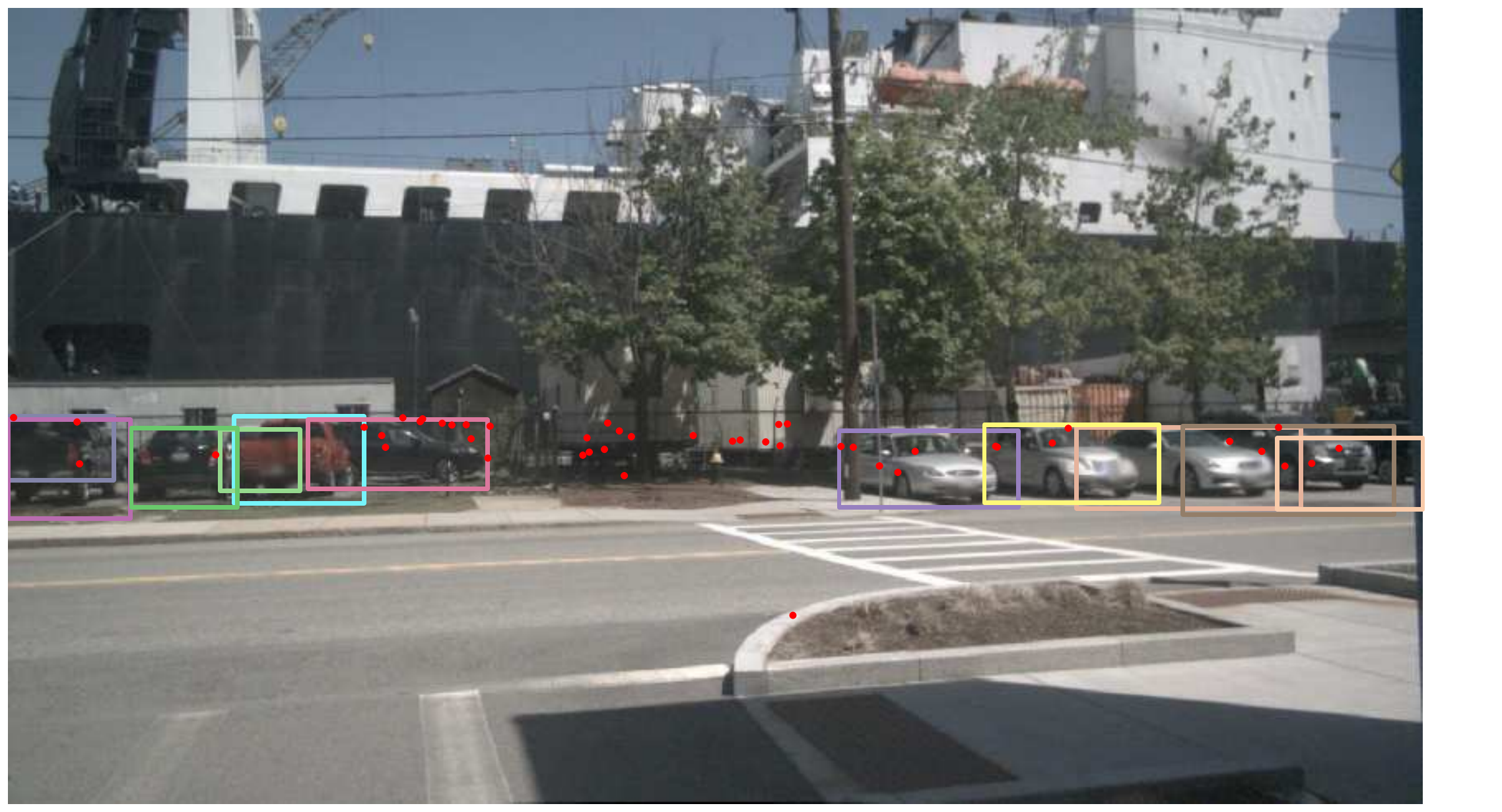}
  \end{minipage}
  \begin{minipage}{0.24\linewidth}
    \includegraphics[width=0.98\textwidth,trim={0 0.2cm 1.2cm 0},clip]{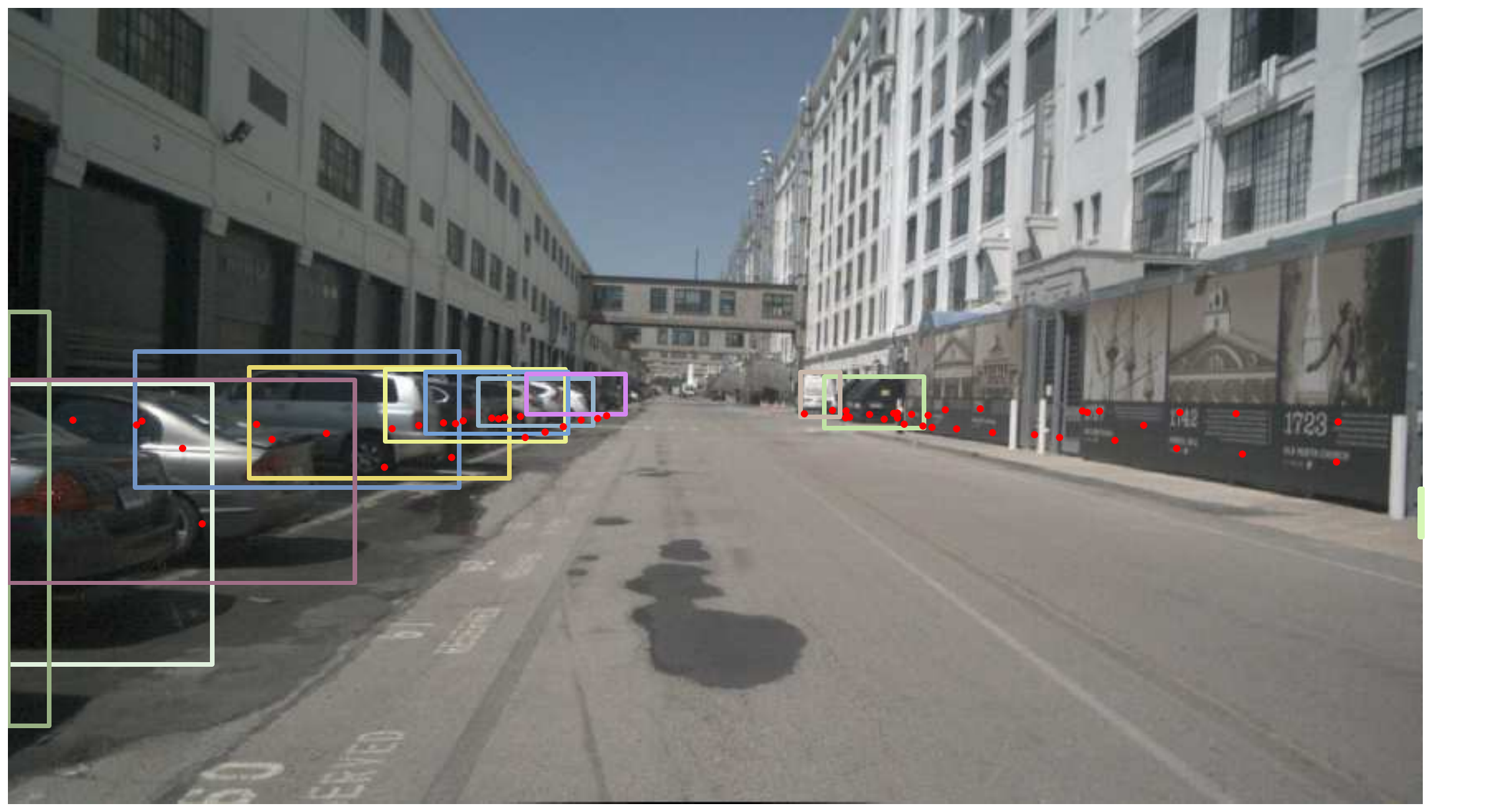}
  \end{minipage}
  \begin{minipage}{0.24\linewidth}
    \includegraphics[width=0.98\textwidth,trim={0 0.2cm 1.2cm 0},clip]{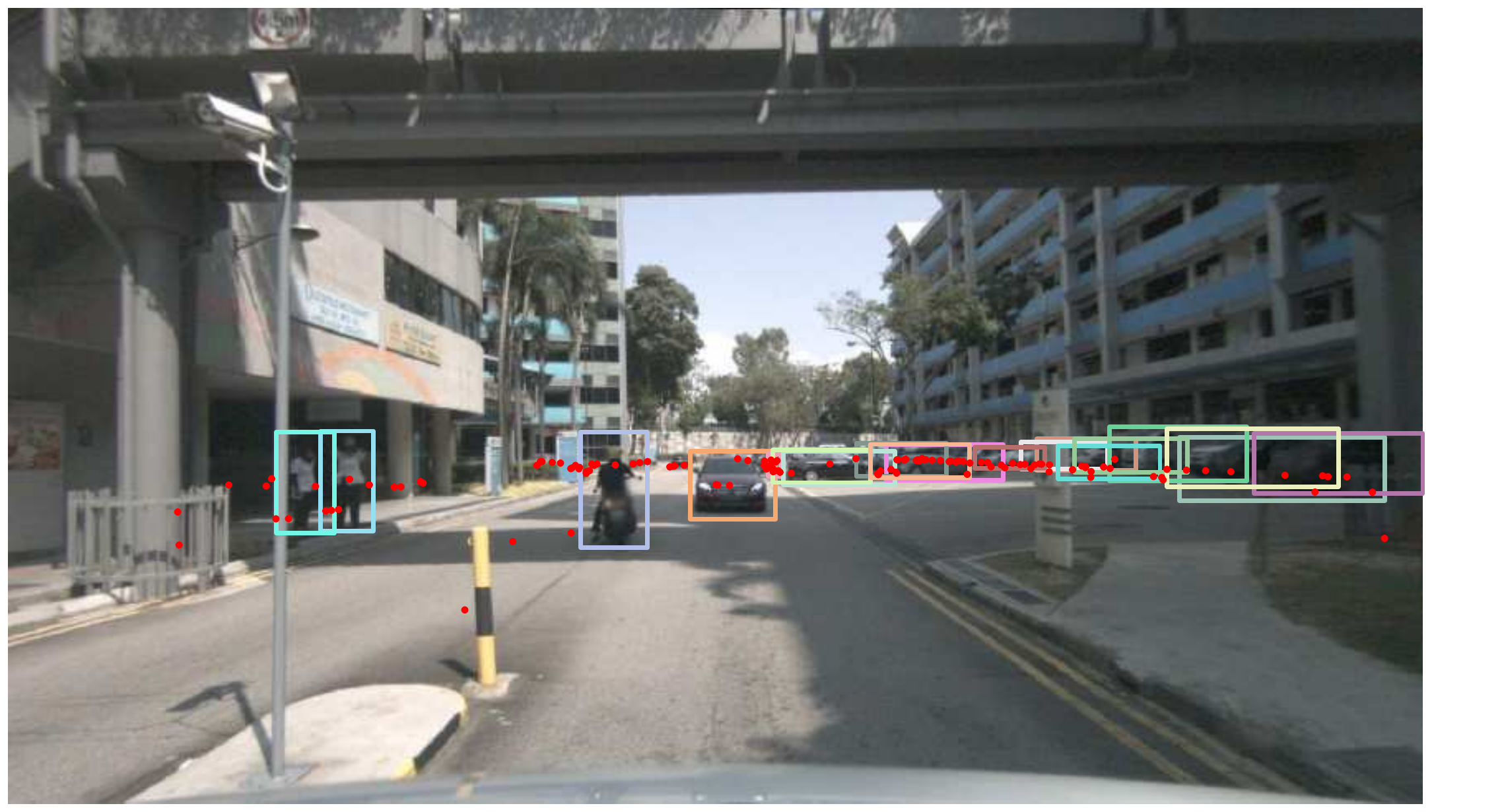}
  \end{minipage}\\
  \begin{minipage}{0.24\linewidth}
    \includegraphics[width=0.98\textwidth]{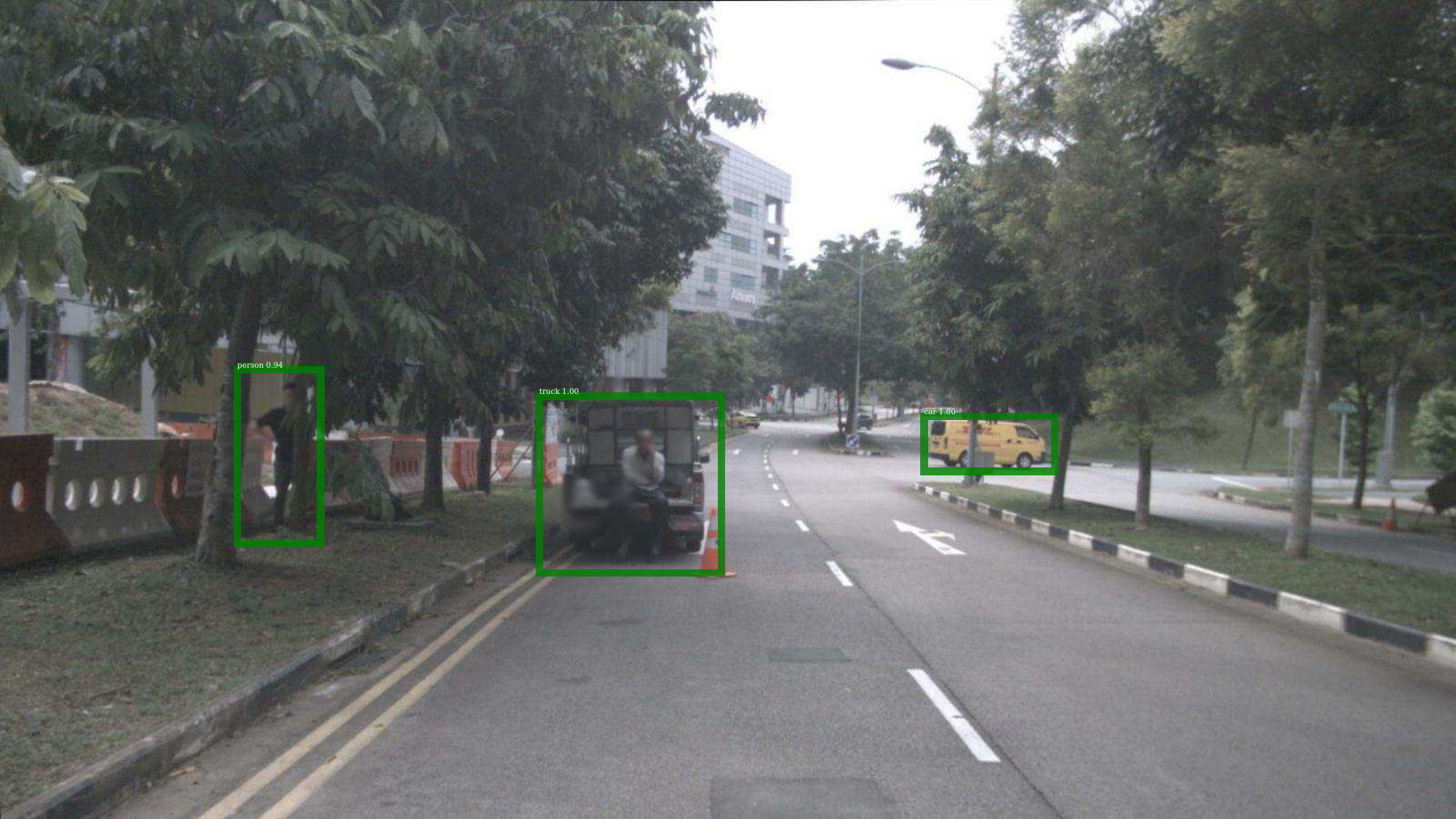}
  \end{minipage}
  \begin{minipage}{0.24\linewidth}
    \includegraphics[width=0.98\textwidth]{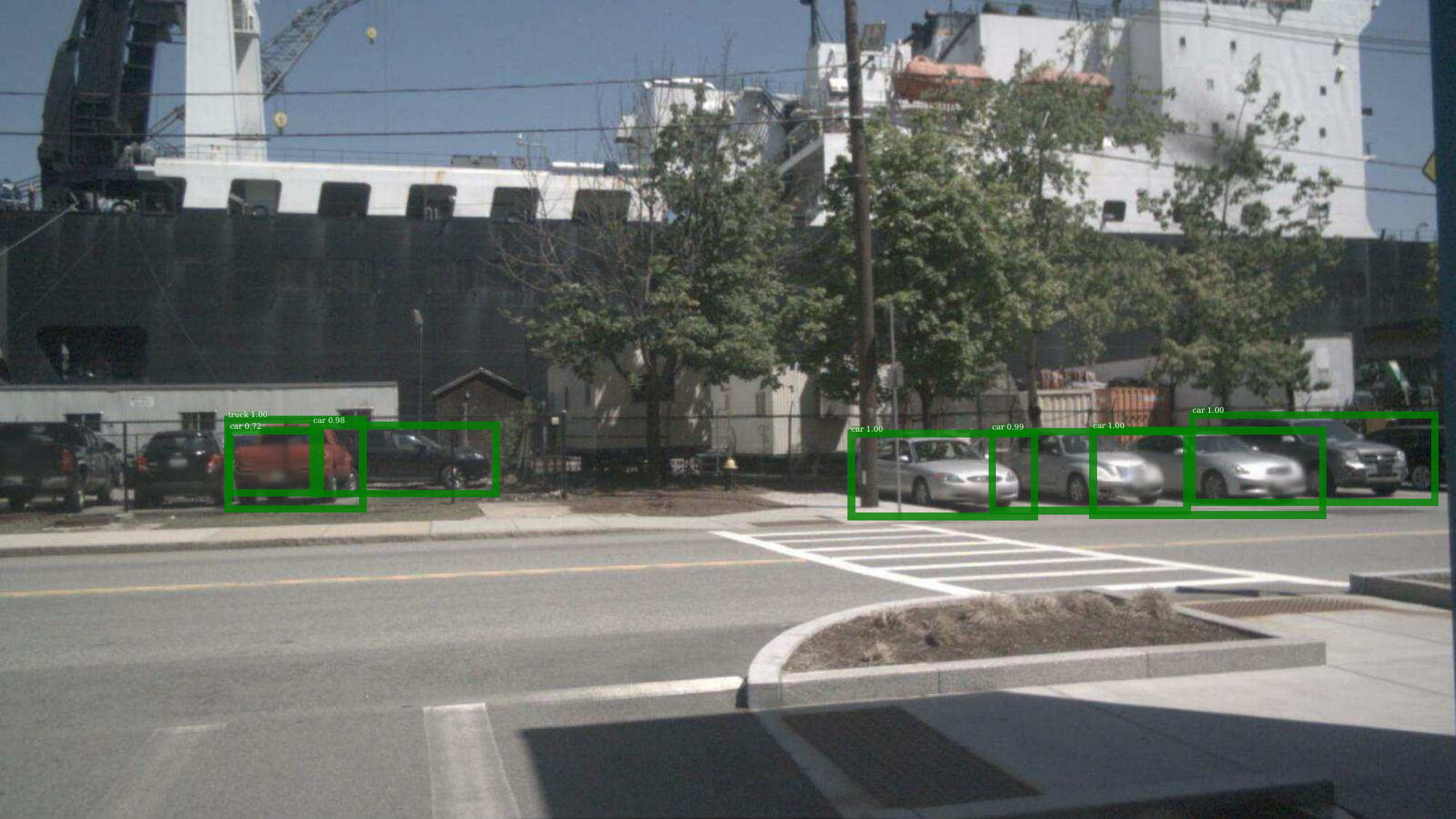}
  \end{minipage}
  \begin{minipage}{0.24\linewidth}
    \includegraphics[width=0.98\textwidth]{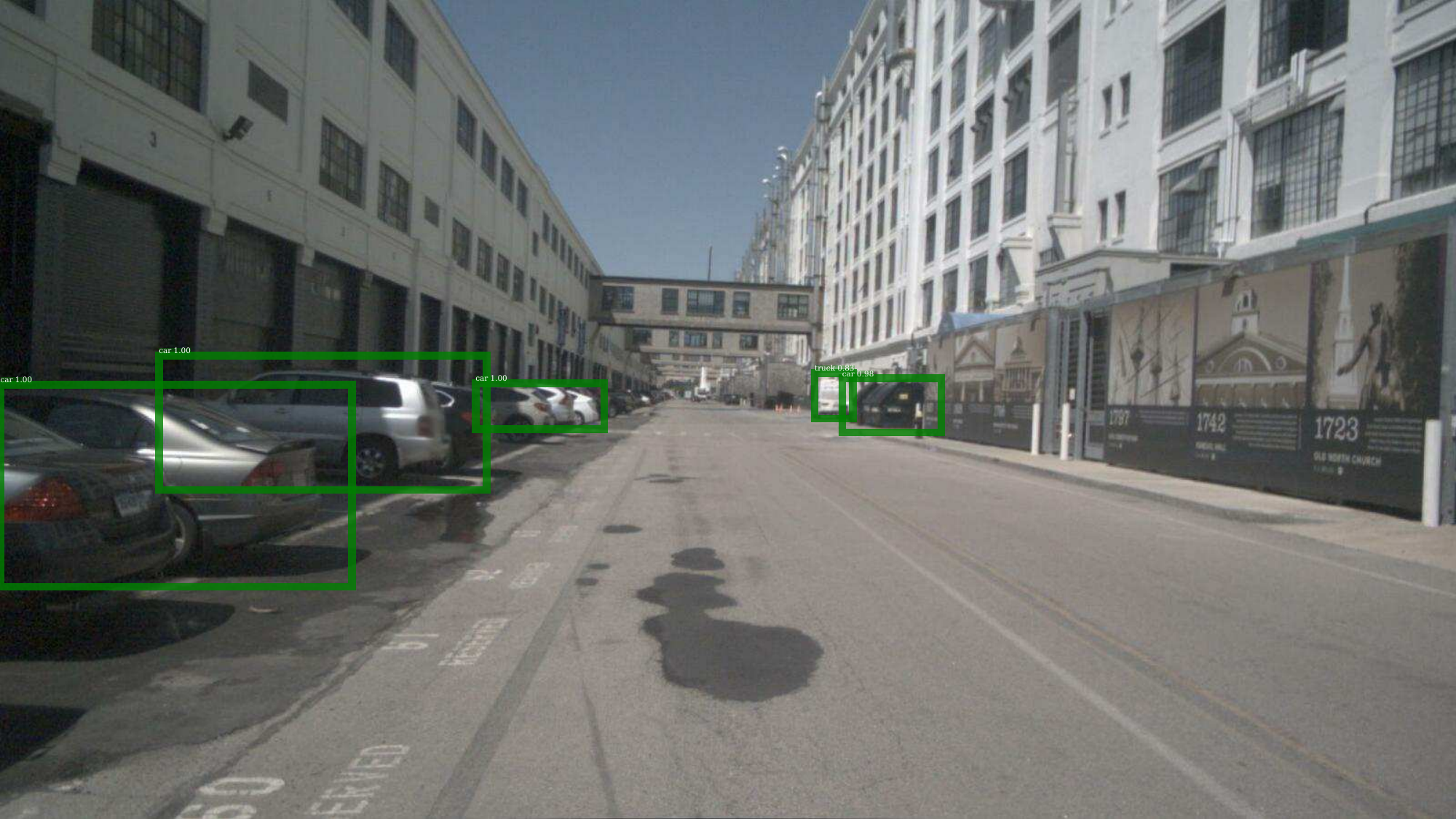}
  \end{minipage}
  \begin{minipage}{0.24\linewidth}
    \centering
    \includegraphics[width=0.98\textwidth]{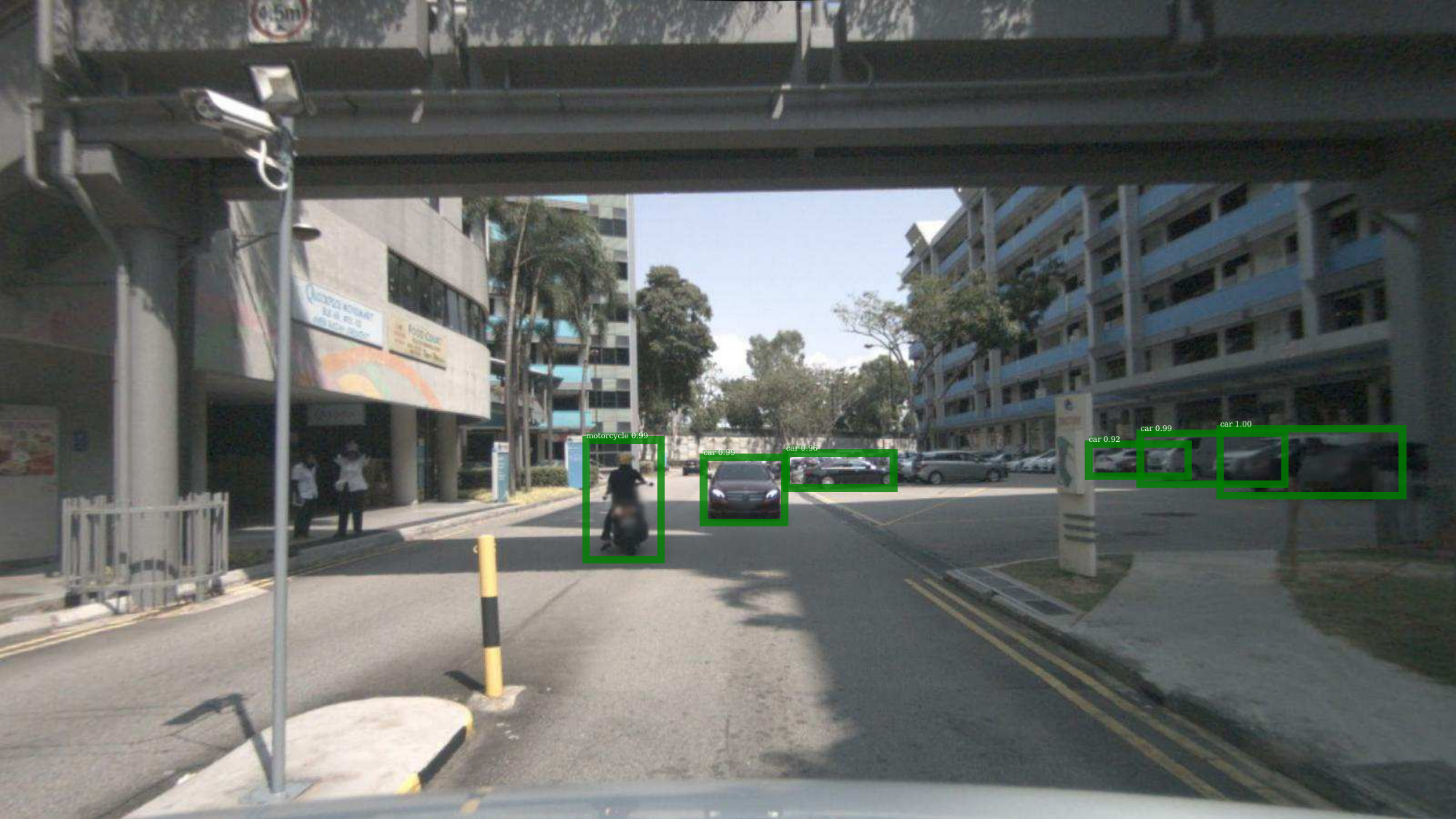}
  \end{minipage}\\
  \begin{minipage}{0.24\linewidth}
    \includegraphics[width=0.98\textwidth]{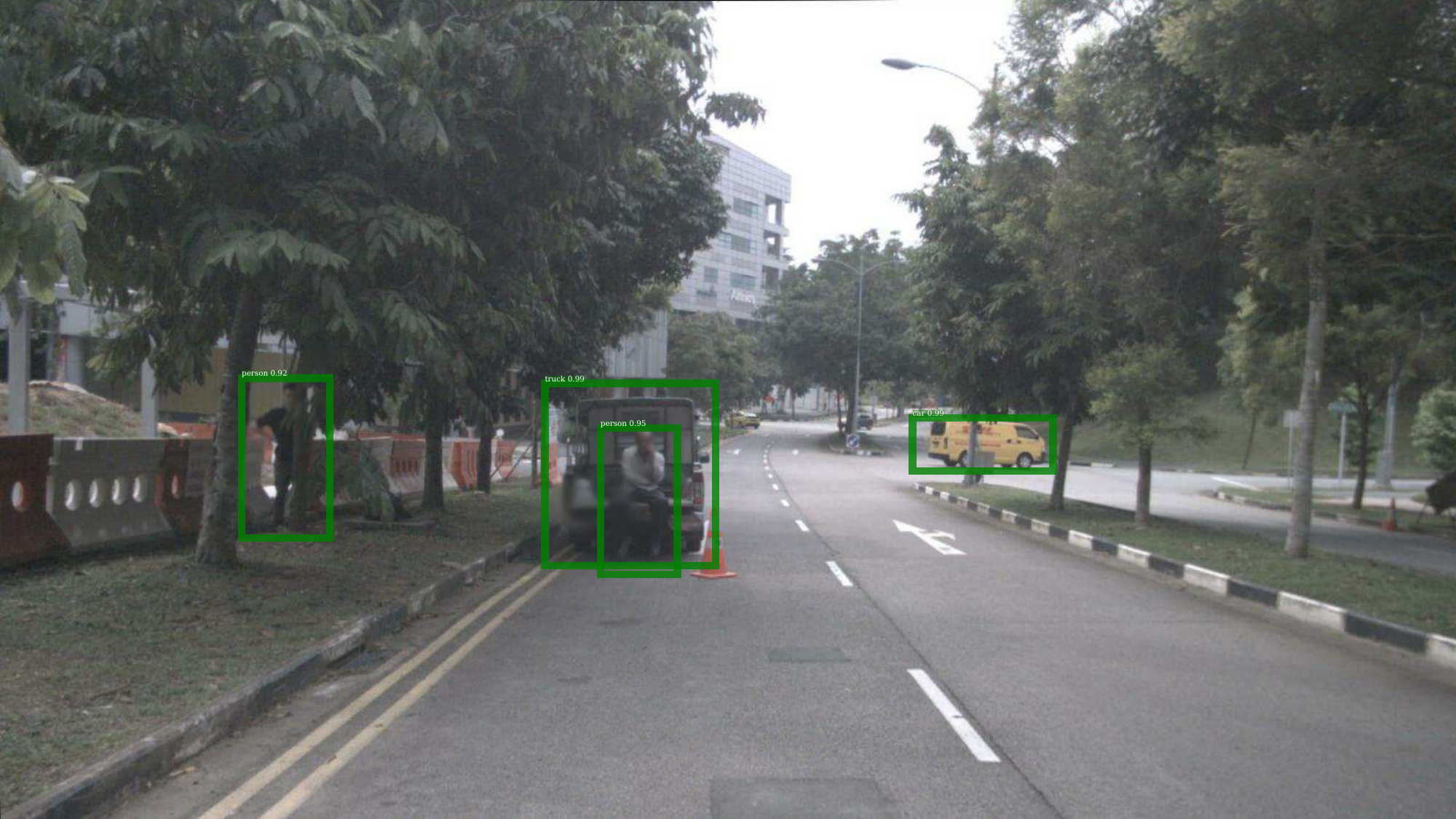}
  \end{minipage}
  \begin{minipage}{0.24\linewidth}
    \includegraphics[width=0.98\textwidth]{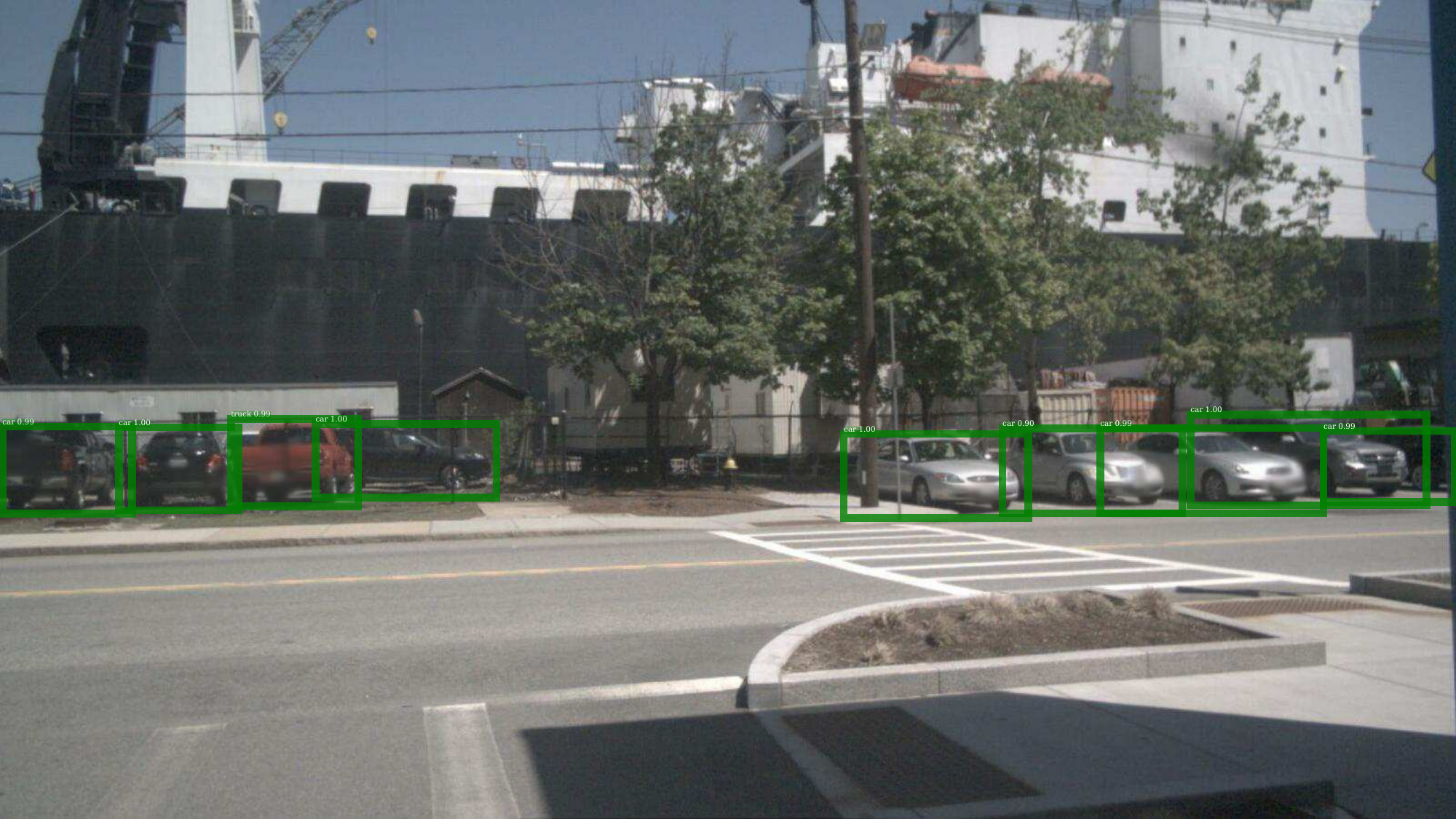}
  \end{minipage}
  \begin{minipage}{0.24\linewidth}
    \includegraphics[width=0.98\textwidth]{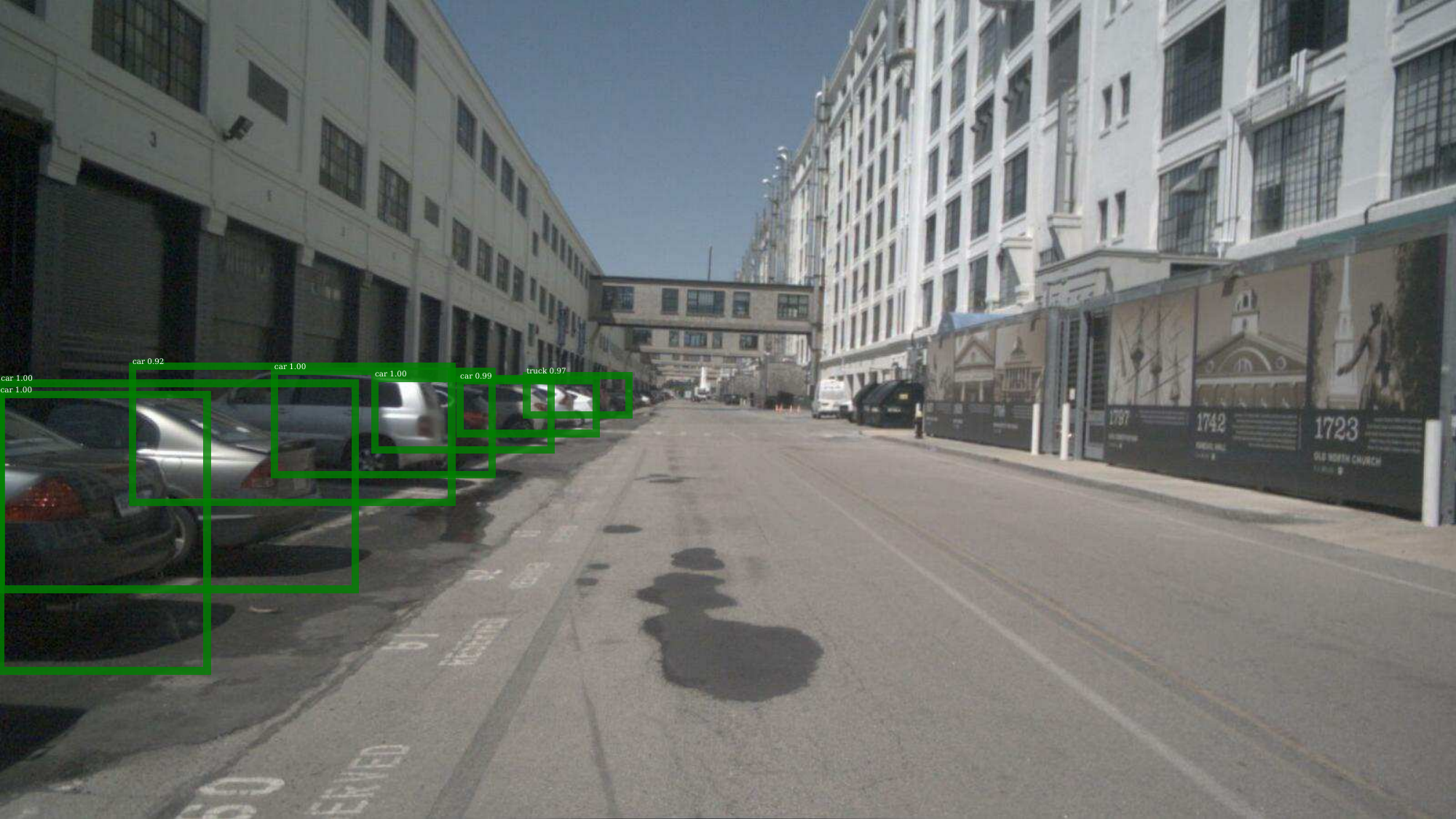}
  \end{minipage}
  \begin{minipage}{0.24\linewidth}
    \includegraphics[width=0.98\textwidth]{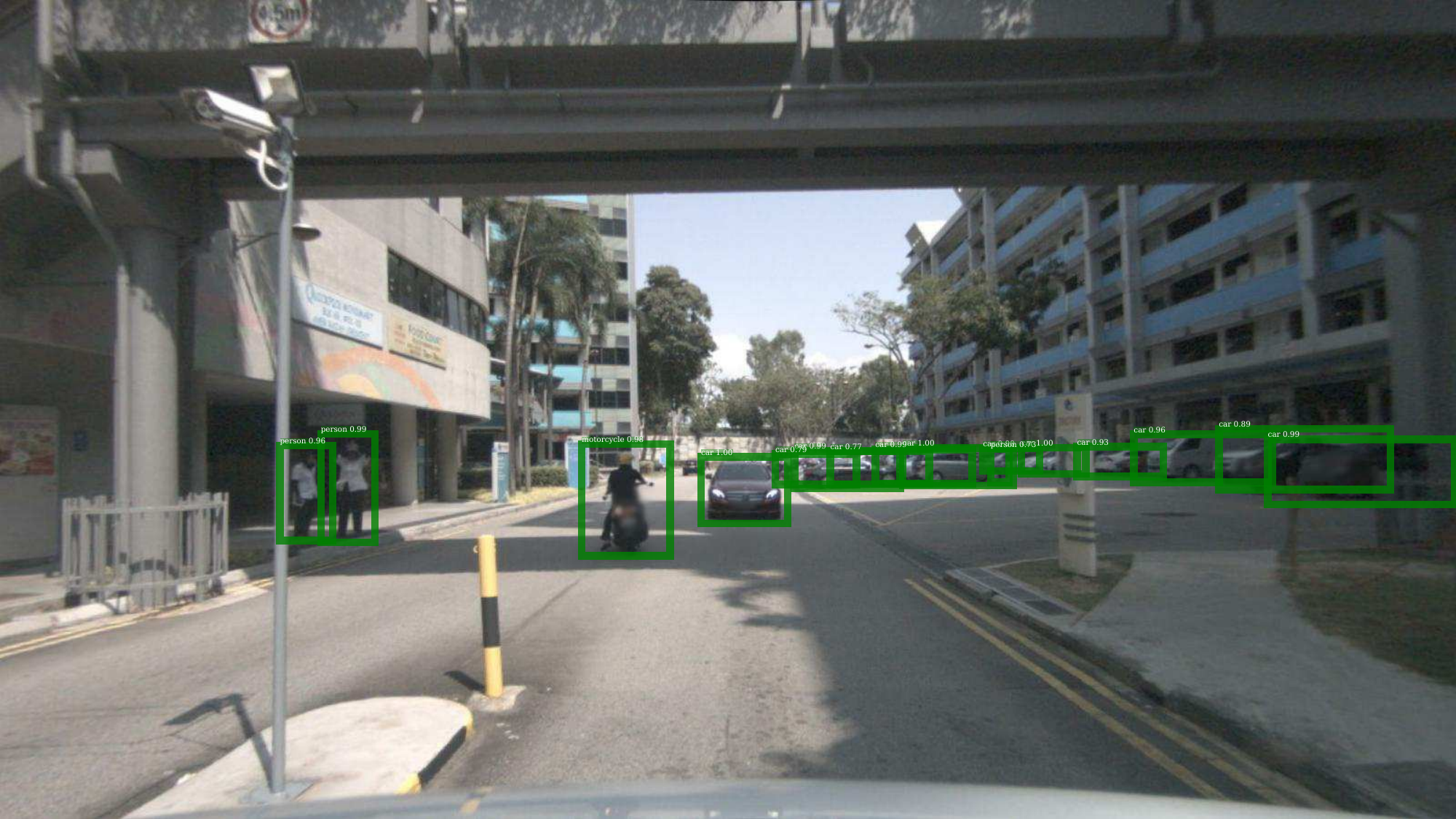}
  \end{minipage}
  \caption{Detection results. Top row: ground truth, middle row: Selective Search, bottom row: RRPN}
  \label{fig:detectionResults}
\end{figure*}

\subsection{Dataset}
To evaluate the proposed RPN, we use the recently released 
\dataset dataset. \dataset is a publicly available large-scale dataset for autonomous 
driving, featuring a full sensor suite including Radars, cameras, LIDAR and GPS 
units. Having 3D bounding boxes for 25 object classes and 1.3M Radar 
sweeps, \dataset is the first large-scale dataset to publicly provide 
synchronized and annotated camera and Radar data collected in highly 
challenging driving situations. To use this dataset in our application, 
we have converted all 3D bounding boxes to 2D 
and also merged some of the similar classes, such as child, adult and police officer. The 
classes used for our experiments are Car, Truck, Person, Motorcycle, Bicycle and Bus.

The \dataset dataset includes images from 6 different cameras in the front, sides and 
back of the vehicle. The Radar detections are obtained from four corner Radars and one 
front Radar. We use two subsets of the samples available in the dataset for our 
experiments. The first subset contains data from the front camera and front Radar only,
with 23k samples. We refer to this subset as \datasetF. The second subset contains data from 
the rear camera and two rear Radars, in addition to all the samples from \datasetF. This 
subset has 45k images and we call it \datasetFB.
Since front Radars usually have a longer range compared to the corner Radars, 
\datasetF gives us more accurate detections for objects far away from the vehicle. 
On the other hand, \datasetFB includes samples from the rear camera and Radar that 
are more challenging for our network. We further 
split each dataset with a ratio of 0.85-0.15 for training and testing, respectively.

\subsection{Implementation Details}
We use the RRPN to propose ROIs for a Fast R-CNN object detection network. 
Two different backbone networks have been used with 
Fast R-CNN: the original ResNet-101 network \cite{He_2016} hereafter 
called R101 and 
the ResNeXt-101 \cite{Xie_2017}, an improved version of ResNet, 
hereafter called X101. In the training stage, we start from a model pre-trained 
on the COCO dataset and further fine-tune it on \datasetF and 
\datasetFB . We compare the results of detection using RRPN proposals with that 
of the Selective Search algorithm \cite{ss2013}, 
which uses a variety of complementary image partitionings to find objects in images. In 
both RRPN and Selective Search, we limit the number of object proposals to 2000 per image.

The evaluation metrics used in our experiments are the same metrics used in 
the COCO dataset \cite{Lin_2014}, namely 
mean Average Precision (AP) and mean Average Recall (AR). 
We also report the AP calculated 
with $0.5$ and $0.75$ IOU, as well as AR for small, medium and large objects areas.
\subsection{Results}
The Fast R-CNN object detection results for the two RPN networks on \datasetF and 
\datasetFB datasets are shown in Table \ref{table:res}. 
According to these results, RRPN is outperforming Selective Search in almost all metrics.
Table \ref{table:clss} shows the per-class AP results for the 
\datasetF and \datasetFB datasets, respectively. For the \datasetF dataset, RRPN outperforms 
Selective Search in the Person, Motorcycle and Bicycle classes with a wide margin, while 
following Selective Search closely in other classes. For the \datasetFB dataset, RRPN 
outperforms Selective Search in all classes except for the Bus class.

Figure \ref{fig:detectionResults} shows selected examples of the object detection results, with the first row showing the ground truth and mapped Radar detections. The next two rows are the detected bounding boxes using the region proposals from 
Selective Search and RRPN respectively. According to these figures, RRPN has been very 
successful in proposing accurate bounding boxes even under hard circumstances such as 
object occlusion and overlap.
In our experiments, RRPN was able to generate proposals for anywhere between 70 to 90 
images per second, depending on the number of Radar detections, while Selective Search took between 2-7 seconds per image.

\begin{table}[t!]
\scriptsize
\centering
\caption{Detection results for the  \datasetF and \datasetFB datasets}
\setlength\tabcolsep{3pt}
\begin{tabular}{l|c|c|c|c|c|c|c}
    method & AP & AP50 & AP75 & AR & ARs & ARm & ARl\\
    \hline\hline
    SS + X101 - F & 0.368 & 0.543 & 0.406 & 0.407 & 0.000 & 0.277 & 0.574\\
    SS + R101 - F & 0.418 & 0.628 & 0.450 & 0.464 & 0.001 & 0.372 & 0.316\\
    RRPN + X101 - F & 0.419 & \textbf{0.652} & 0.463 & 0.478 & \textbf{0.041} & 0.406 & 0.573\\
    RRPN + R101 - F & \textbf{0.430} & 0.649 & \textbf{0.485} & \textbf{0.486} & 0.040 & \textbf{0.412} & \textbf{0.582}\\
    \hline
    SS + X101 - FB& 0.332 & 0.545 & 0.352 & 0.382 & 0.001 & 0.291 & 0.585\\
    SS + R101 - FB& 0.336 & 0.548 & 0.357 & 0.385 & 0.001 & 0.291 & \textbf{0.591}\\
    RRPN + X101 - FB& 0.354 & \textbf{0.592} & 0.369 & 0.420 & 0.202 & \textbf{0.391} & 0.510\\
    RRPN + R101 - FB& \textbf{0.355} & 0.590 & \textbf{0.370} & \textbf{0.421} & \textbf{0.211} & \textbf{0.391} & 0.514\\
\end{tabular}
\label{table:res}
\end{table}

\begin{table}[t!]
\scriptsize
\centering
\caption{Per-class AP for the \datasetF and \datasetFB datasets}
\setlength\tabcolsep{3pt}
\begin{tabular}{l|c|c|c|c|c|c}
    method & Car & Truck & Person & Motorcycle & Bicycle & Bus \\
    \hline\hline
    SS + X101 - F & 0.424 & 0.509 & 0.117 & 0.288 & 0.190 & 0.680\\
    SS + R101 - F& \textbf{0.472} & \textbf{0.545} & 0.155 & 0.354 & 0.241 & \textbf{0.722}\\
    RRPN + X101 - F & 0.428 & 0.501 & 0.212 & 0.407 & 0.304 & 0.660\\
    RRPN + R101 - F& 0.442 & 0.516 & \textbf{0.220} & \textbf{0.434} & \textbf{0.306} & 0.664\\
    \hline
    SS + X101 - FB& 0.390 & 0.415 & 0.122 & 0.292 & 0.179 & 0.592\\
    SS + R101 - FB& 0.392 & 0.420 & 0.121 & 0.291 & 0.191 & \textbf{0.600}\\
    RRPN + X101 - FB& 0.414 & \textbf{0.449} & \textbf{0.174} & 0.294 & \textbf{0.215} & 0.579\\
    RRPN + R101 - FB& \textbf{0.418} & 0.447 & 0.171 & \textbf{0.305} & 0.214 & 0.572\\
\end{tabular}
\label{table:clss}
\end{table}

\section{Conclusion}
\label{sec:conclusion}
We presented RRPN, a real-time region proposal network for object 
detection in autonomous driving applications. By only relying on Radar detections to 
propose ROIs, our method is extremely fast while at the same time achieving a higher 
precision and recall compared to the Selective Search algorithm. Additionally, RRPN 
inherently performs as a sensor fusion algorithm, fusing the data obtained 
from Radars with vision data to obtain faster and more accurate detections.
We evaluated RRPN on the \dataset dataset and compared the 
results to the Selective Search algorithm. Our experiments show RRPN operates more 
than 100x faster than the Selective Search algorithm, while resulting 
in better detection average precision and recall.

\bibliographystyle{IEEEbib}
\bibliography{refs}

\begin{thebibliography}{10}

\bibitem{caesar2019nuscenes}
Holger Caesar, Varun Bankiti, Alex~H. Lang, Sourabh Vora, Venice~Erin Liong,
  Qiang Xu, Anush Krishnan, Yu~Pan, Giancarlo Baldan, and Oscar Beijbom,
\newblock ``nuscenes: A multimodal dataset for autonomous driving,'' 2019.

\bibitem{Girshick2015}
Ross Girshick,
\newblock ``Fast {R-CNN},''
\newblock {\em 2015 IEEE International Conference on Computer Vision (ICCV)},
  Dec 2015.

\bibitem{ss2013}
Jasper R.~R. Uijlings, Koen E.~A. van~de Sande, Theo Gevers, and Arnold W.~M.
  Smeulders,
\newblock ``{Selective Search for Object Recognition},''
\newblock {\em International Journal of Computer Vision}, vol. 104, no. 2, pp.
  154--171, sep 2013.

\bibitem{Grimes1974}
Dale~M. Grimes and Trevor~Owen Jones,
\newblock ``{Automotive Radar: A Brief Review},''
\newblock {\em Proceedings of the IEEE}, vol. 62, no. 6, pp. 804--822, 1974.

\bibitem{rfcn2016r}
Jifeng Dai, Yi~Li, Kaiming He, and Jian Sun,
\newblock ``R-fcn: Object detection via region-based fully convolutional
  networks,''
\newblock in {\em Advances in neural information processing systems}, 2016, pp.
  379--387.

\bibitem{ren2015faster}
Shaoqing Ren, Kaiming He, Ross Girshick, and Jian Sun,
\newblock ``{Faster R-CNN: Towards Real-Time Object Detection with Region
  Proposal Networks},''
\newblock in {\em Neural Information Processing Systems (NIPS)}, jun 2015.

\bibitem{liu2016ssd}
Wei Liu, Dragomir Anguelov, Dumitru Erhan, Christian Szegedy, Scott Reed,
  Cheng-Yang Fu, and Alexander~C. Berg,
\newblock ``{SSD: Single Shot MultiBox Detector},''
\newblock in {\em European Conference on Computer Vision}. dec 2016.

\bibitem{soviany2018optimizing}
Petru Soviany and Radu~Tudor Ionescu,
\newblock ``Optimizing the trade-off between single-stage and two-stage object
  detectors using image difficulty prediction,''
\newblock {\em arXiv preprint arXiv:1803.08707}, 2018.

\bibitem{redmon2016YOLO}
Joseph Redmon, Santosh Divvala, Ross Girshick, and Ali Farhadi,
\newblock ``{You Only Look Once: Unified, Real-Time Object Detection},''
\newblock in {\em 2016 IEEE Conference on Computer Vision and Pattern
  Recognition (CVPR)}. jun 2016, pp. 779--788, IEEE.

\bibitem{Gibson1994}
R~E Gibson, D~L Hall, and J~a Stover,
\newblock ``{An autonomous fuzzy logic architecture for multisensor data
  fusion},''
\newblock {\em Proceedings of 1994 IEEE International Conference on MFI 94
  Multisensor Fusion and Integration for Intelligent Systems}, vol. 43, no. 3,
  pp. 403--410, 1994.

\bibitem{Miyahara2006}
Shunji Miyahara, Jerry Sielagoski, Anatoli Koulinitch, and Faroog Ibrahim,
\newblock ``Target tracking by a single camera based on range-window algorithm
  and pattern matching,''
\newblock in {\em SAE Technical Paper}. 04 2006, SAE International.

\bibitem{Ji2008}
Zhengping Ji and Danil Prokhorov,
\newblock ``{Radar-vision fusion for object classification},''
\newblock {\em Proceedings of the 11th International Conference on Information
  Fusion, FUSION 2008}, vol. 2, pp. 265--271, 2008.

\bibitem{Premebida2009}
Cristiano Premebida, Oswaldo Ludwig, and Urbano Nunes,
\newblock ``Lidar and vision-based pedestrian detection system,''
\newblock {\em Journal of Field Robotics}, vol. 26, no. 9, pp. 696--711, 2009.

\bibitem{Cho2014}
Hyunggi Cho, Young~Woo Seo, B.~V.K.Vijaya Kumar, and Ragunathan~Raj Rajkumar,
\newblock ``{A multi-sensor fusion system for moving object detection and
  tracking in urban driving environments},''
\newblock {\em Proceedings - IEEE International Conference on Robotics and
  Automation}, pp. 1836--1843, 2014.

\bibitem{edge2014}
C~Lawrence Zitnick and Piotr Doll{\'a}r,
\newblock ``Edge boxes: Locating object proposals from edges,''
\newblock in {\em European conference on computer vision}. Springer, 2014, pp.
  391--405.

\bibitem{He_2016}
Kaiming He, Xiangyu Zhang, Shaoqing Ren, and Jian Sun,
\newblock ``Deep residual learning for image recognition,''
\newblock {\em 2016 IEEE Conference on Computer Vision and Pattern Recognition
  (CVPR)}, Jun 2016.

\bibitem{Xie_2017}
Saining Xie, Ross Girshick, Piotr Dollar, Zhuowen Tu, and Kaiming He,
\newblock ``Aggregated residual transformations for deep neural networks,''
\newblock {\em 2017 IEEE Conference on Computer Vision and Pattern Recognition
  (CVPR)}, Jul 2017.

\bibitem{Lin_2014}
Tsung-Yi Lin, Michael Maire, Serge Belongie, James Hays, Pietro Perona, Deva
  Ramanan, Piotr Dollár, and C.~Lawrence Zitnick,
\newblock ``Microsoft coco: Common objects in context,''
\newblock {\em Lecture Notes in Computer Science}, p. 740–755, 2014.

\end{thebibliography}

\end{document}